\documentclass[sigconf,pbalance,nonacm]{acmart}

\usepackage{booktabs}
\usepackage{multirow}
\usepackage{amsmath}
\usepackage{array}
\usepackage{xspace}
\usepackage{algorithm}
\usepackage{algpseudocode}
\usepackage{subcaption}

\newcommand{\method}{SCOPE\xspace}
\newcommand{\cache}{\textsc{Cache}\xspace}
\newcommand{\predict}{\textsc{Predict}\xspace}
\newcommand{\recompute}{\textsc{Recompute}\xspace}
\newcommand{\deltadit}{$\Delta$-DiT\xspace}


\begin{document} 

%% Paper Title
\title[SCOPE]{Not All Frames Deserve Full Computation: Accelerating Autoregressive Video Generation via Selective Computation and Predictive Extrapolation}

%% Consolidated author line and affiliations for the preprint.
\author{\fontsize{9}{10}\selectfont Hanshuai Cui$^{2,1}$, Zhiqing Tang$^{1}$, Zhi Yao$^{2,1}$, Fanshuai Meng$^{1}$, Weijia Jia$^{1}$, Wei Zhao$^{3}$}
\affiliation{%
  \institution{$^{1}$Institute of Artificial Intelligence and Future Networks, Beijing Normal University\city{Zhuhai}\country{China}}
}
\affiliation{%
  \institution{$^{2}$School of Artificial Intelligence, Beijing Normal University\city{Beijing}\country{China}}
}
\affiliation{%
  \institution{$^{3}$Shenzhen University of Advanced Technology\city{Shenzhen}\country{China}}
}
%% Short author list for page header
\renewcommand{\shortauthors}{Cui et al.}

%% Abstract
\begin{abstract}
Autoregressive (AR) video diffusion models enable long-form video generation but remain expensive due to repeated multi-step denoising. Existing training-free acceleration methods rely on binary cache-or-recompute decisions, overlooking intermediate cases where direct reuse is too coarse yet full recomputation is unnecessary. Moreover, asynchronous AR schedules assign different noise levels to co-generated frames, yet existing methods process the entire valid interval uniformly. To address these AR-specific inefficiencies, we present SCOPE, a training-free framework for efficient AR video diffusion. SCOPE introduces a tri-modal scheduler over cache, predict, and recompute, where prediction via noise-level Taylor extrapolation fills the gap between reuse and recomputation with explicit stability controls backed by error propagation analysis. It further introduces selective computation that restricts execution to the active frame interval. On MAGI-1 and SkyReels-V2, SCOPE achieves up to 4.73x speedup while maintaining quality comparable to the original output, outperforming all training-free baselines. Our code is publicly available at \url{https://github.com/hsc113/SCOPE}.
\end{abstract}

\maketitle

%% Main Content
\section{Introduction}

Autoregressive (AR) video diffusion has emerged as a leading paradigm for long-form video generation~\cite{ai2025magi1,chen2025skyreelsv2}. It extends videos causally, chunk by chunk or frame by frame, while conditioning on previously generated content. Built upon diffusion transformers (DiTs)~\cite{peebles2023dit,esser2024sd3} and flow-matching objectives~\cite{lipman2023flowmatching,liu2023rectifiedflow}, often accelerated by efficient attention mechanisms~\cite{dao2022flashattention,dao2024flashattention2,shazeer2019mqa}, these models achieve impressive quality on videos spanning hundreds of frames~\cite{brooks2024sora,polyak2024moviegen,bartal2024lumiere}. However, a fundamental inference bottleneck persists, where every AR unit still performs multi-step diffusion integration, so generating long videos requires many expensive transformer forwards.

\begin{figure}[t]
  \centering
  \includegraphics[width=\columnwidth]{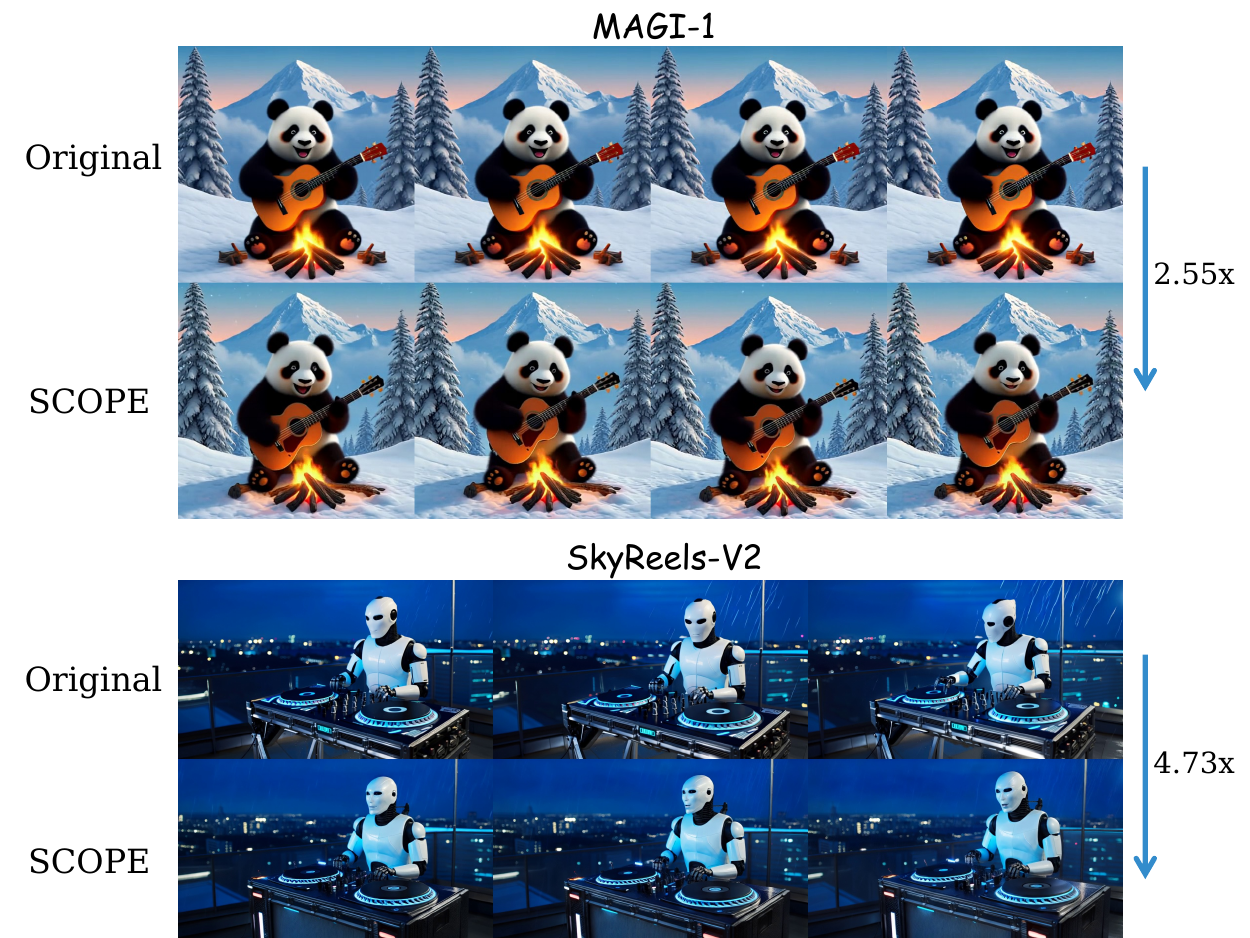}
  \caption{SCOPE achieves large speedup while preserving near-original quality on both MAGI-1 and SkyReels-V2.}
  \Description{A teaser figure showing SCOPE's speedup over the original pipeline on MAGI-1 and SkyReels-V2.}
  \label{fig:teaser_speedup}
\end{figure}

This cost has motivated a growing body of training-free acceleration methods, spanning block caching~\cite{wimbauer2024cacheme}, layer caching~\cite{ma2024learningtocache}, universal feature reuse~\cite{so2024frdiff}, and more recent reuse-oriented schemes~\cite{liu2025teacache,chen2025deltadit,liu2025speca}. Most existing methods primarily exploit step-level or block-level similarity in standard diffusion, while FlowCache~\cite{ma2026flowcache} targets AR video generation with chunk-aware adaptive caching. However, these methods still rely on a binary cache-or-recompute decision. This leaves a structural gap, as some steps are too inaccurate for direct reuse but too smooth to justify a full forward pass. When forced into binary decisions, either quality degrades from excessive reuse or speedup saturates from conservative recomputation. Forecast-style extrapolation is also not new in itself, as TaylorSeers~\cite{liu2025taylorseers} already predicts diffusion features via Taylor expansion. Our goal is making it usable inside AR video diffusion, where prediction must be embedded into a stable step scheduler and paired with the frame-wise activity structure of asynchronous AR denoising.

Beyond the binary cache-or-recompute gap, AR video diffusion also exhibits a second source of inefficiency. Recent causal video diffusion models expose richer inference structure than standard synchronous denoising~\cite{chen2024diffusionforcing,chen2025skyreelsv2,ai2025magi1,yin2025causvid}. Under asynchronous AR schedules~\cite{chen2024diffusionforcing,chen2025skyreelsv2}, different frames within the same iteration can reside at different denoising stages, forming a staircase-like step matrix. Only a subset of frames is actively evolving at any given iteration. Existing acceleration methods, however, still process the entire valid interval uniformly, wasting computation on nearly converged frames at the tail and on not-yet-started frames at the head. This spatial redundancy becomes especially severe in long-form generation. The valid interval may span dozens of frames even though only a small active subset requires meaningful updates~\cite{ai2025magi1,yin2025causvid}. AR inference therefore suffers from two distinct forms of waste, temporal redundancy across denoising steps and spatial redundancy within the current valid interval.

Building an effective acceleration method for AR diffusion faces two challenges. \textbf{(C1)} How to bridge the gap between direct reuse and full recomputation under AR causal denoising? Approximation errors accumulate over consecutive non-recompute steps and propagate into later causal extensions, so an intermediate prediction mechanism must remain stable across an entire AR rollout. \textbf{(C2)} How to reduce redundant computation inside asynchronous AR schedules? Because the active denoising frontier moves over time and only a subset of the full valid interval requires meaningful updates, selective execution must efficiently focus computation on the active frame interval.

We propose \method (\textbf{S}elective \textbf{CO}mputation with \textbf{P}redictive \textbf{E}xtrapolation), a training-free acceleration framework for AR video diffusion models. First, \method introduces a \predict mode that fills the gap between direct reuse and full recomputation by forecasting the velocity via local Taylor extrapolation along the denoising trajectory, producing a tri-modal policy over \cache, \predict, and \recompute. To maintain prediction stability over long AR rollouts, the scheduler incorporates several stability controls to improve robustness. Second, \method introduces selective computation, which restricts execution to the active frame interval in asynchronous AR schedules, reducing per-iteration cost without altering the AR schedule itself. As a training-free method, the integration overhead of \method is minimal. It requires only a lightweight pre-forward signal for risk estimation, access to recent denoising history for prediction, and support for restricting computation to the active frame interval. We validate \method on MAGI-1 and SkyReels-V2, achieving up to $2.55\times$ speedup on MAGI-1 and up to $4.73\times$ speedup on SkyReels-V2. As shown in Figure~\ref{fig:teaser_speedup}, \method reduces both temporal and spatial redundancy while keeping quality close to the Original pipeline. Our main contributions are:

\begin{itemize}
  \item We formulate AR denoising control as a tri-modal scheduler over \cache, \predict, and \recompute, introducing prediction as an explicit intermediate mode between reuse and recomputation with stability controls for long AR rollouts.
  \item We propose selective computation for asynchronous AR schedules, restricting execution to the active frame interval to reduce frame-level redundant computation while remaining fully compatible with the \predict branch.
  \item We validate \method on two representative AR video diffusion models, MAGI-1 and SkyReels-V2, demonstrating that the tri-modal scheduler and selective computation together yield the strongest speed-quality tradeoff among all training-free baselines.
\end{itemize}

\section{Related Work}

\subsection{Autoregressive Video Diffusion Models}

Video generation has advanced rapidly through diffusion models~\cite{ho2020ddpm,song2021scorebased,nichol2021improved,dhariwal2021beatgans,rombach2022latentdiffusion,ho2022videodiffusion} and their transformer-based variants~\cite{vaswani2017attention,dosovitskiy2021vit,peebles2023dit,bao2023uvit}. Recent models produce high-quality videos via one-shot~\cite{blattmann2023svd,guo2024animatediff,wang2023lavie,he2023lvdm,ho2022imagenvideo,singer2023makeavideo,bartal2024lumiere,zheng2024opensora,yang2024cogvideox} or autoregressive generation~\cite{hong2023cogvideo}. Among AR approaches, SkyReels-V2~\cite{chen2025skyreelsv2} adopts a diffusion forcing~\cite{chen2024diffusionforcing} pipeline where frames carry different noise levels, producing a staircase-like step matrix. MAGI-1~\cite{ai2025magi1} uses a chunk-causal design where fixed-length chunks are denoised holistically. Both models couple autoregressive conditioning with flow-matching-based denoising~\cite{lipman2023flowmatching,liu2023rectifiedflow,tong2024cfm}, making their cost scale with video length and denoising depth.

\subsection{Training-Free Acceleration for Diffusion}

Training-free methods reduce diffusion inference cost without fine-tuning or distillation~\cite{salimans2022progressive,song2023consistency,luo2023lcm,liu2024instaflow,yin2024dmd,wu2025individual}, exploiting the smoothness of neighboring denoising states~\cite{song2021ddim}. These methods can be grouped into three broad categories.

\textbf{Block- and layer-level caching.} Cache Me if You Can~\cite{wimbauer2024cacheme} introduces block-level caching for DiTs, and Learning-to-Cache~\cite{ma2024learningtocache} learns layer-level caching policies. $\Delta$-DiT~\cite{chen2025deltadit} exploits stage dependent roles of DiT blocks for block-level reuse. FRDiff~\cite{so2024frdiff} reuses features universally across blocks. These methods target intra-model redundancy but remain binary in their reuse decisions.

\textbf{Attention- and token-level methods.} FasterCache~\cite{lv2024fastercache} exploits redundancy between conditional and unconditional features. PAB~\cite{zhao2024pab} broadcasts attention across timesteps. T-GATE~\cite{pan2024tgate} prunes cross-attention after early steps. Token merging~\cite{bolya2023tome} reduces token counts to lower per-step cost. While effective for reducing per-step FLOPs, these methods do not adapt to the step-level decision of whether to skip computation entirely.

\textbf{Step-level caching and prediction.} AdaCache~\cite{kahatapitiya2024adacache} adapts caching schedules to video content. TeaCache~\cite{liu2025teacache} estimates step similarity from modulated timestep embeddings. TaylorSeers~\cite{liu2025taylorseers} moves beyond binary reuse by forecasting diffusion features via Taylor expansion, but lacks the stability controls needed for long AR rollouts. \method builds on this line by embedding prediction into a tri-modal scheduler with explicit drift control, and by additionally exploiting the frame-wise activity structure unique to asynchronous AR denoising.

\subsection{Acceleration for AR Video Generation}

AR video generation exposes additional structure that generic acceleration methods do not adapt to. FlowCache~\cite{ma2026flowcache} discovers chunk-level heterogeneity and performs chunk-wise adaptive caching, optionally combining reuse with KV compression. However, FlowCache retains a binary cache-or-recompute decision within each chunk, leaving no mechanism for lightweight approximation when neither direct reuse nor full recomputation is appropriate. It also does not exploit the frame-level activity structure of asynchronous schedules, processing the entire valid interval uniformly even when only a subset of frames is actively denoising. Recent work on autoregressive image generation~\cite{li2024mar,tian2024var} and fast AR video diffusion~\cite{yin2025causvid} further motivates efficient inference for causal generative models. \method is complementary to these methods. Rather than modeling chunk heterogeneity or reducing KV cost, \method addresses two complementary gaps: it replaces the binary cache-or-recompute decision with predictive interpolation backed by stability controls for long AR rollouts, and it eliminates redundant computation on inactive frames via dynamic interval shrinking.

\section{Methodology}

\subsection{Preliminaries}

\textbf{Flow-matching denoising.} We consider AR video diffusion models trained under flow-matching~\cite{lipman2023flowmatching,liu2023rectifiedflow}. A noisy latent at noise level $\sigma$ is $x_{\sigma} = (1 - \sigma) x_0 + \sigma \epsilon$, $\epsilon \sim \mathcal{N}(0, I)$. The model parameterizes a velocity field $v_{\theta}(x_{\sigma}, \sigma)$ whose reverse ODE $dx/d\sigma = v_{\theta}$ is integrated by Euler steps:
\begin{equation}
  x_{k-1} = x_k + v_{\theta}(x_k, \sigma_k)\Delta \sigma_k, \qquad \Delta \sigma_k = \sigma_{k-1} - \sigma_k.
  \label{eq:euler-update}
\end{equation}

The high cost of AR video diffusion comes from repeatedly evaluating $v_{\theta}$ across many denoising steps and many AR units.

\textbf{Autoregressive schedules.} AR diffusion models extend a video causally, but the denoising unit may be a synchronized chunk or an asynchronous frame interval~\cite{sun2025ar}. In chunk-synchronized schedules (e.g., MAGI-1), all frames within the AR unit share a common denoising stage, and decisions naturally operate per chunk. The dominant source of waste is repeated unit-level forwards. In asynchronous schedules (e.g., SkyReels-V2), different frames occupy different noise levels within the same iteration, producing frame-wise heterogeneity. Here, decisions operate per iteration or per active frame interval, and the primary waste comes from computing inactive frames inside the valid interval.

\textbf{Local predictability along the denoising trajectory.} TeaCache exploits smoothness across neighboring timesteps through timestep embeddings, and TaylorSeers forecasts future features via Taylor expansion across timesteps~\cite{liu2025teacache,liu2025taylorseers}. Motivated by the same local regularity, we parameterize denoising progression by the continuous noise level $\sigma = g(t)$, where $g$ is the scheduler-specific monotone map used in the implementation to convert the raw scheduler variable $t$ into a noise-level coordinate. We then treat the velocity field as locally smooth along the denoising trajectory. For neighboring states we can write:
\begin{equation}
  v(\sigma + \Delta \sigma) \approx v(\sigma) + \frac{dv}{d\sigma}\Delta \sigma + O(\Delta \sigma^2).
  \label{eq:taylor-intuition}
\end{equation}

\begin{figure}[t]
  \centering
  \subcaptionbox{Feature trajectory\label{fig:binary-gap-curve}}{%
    \includegraphics[width=0.4898\linewidth]{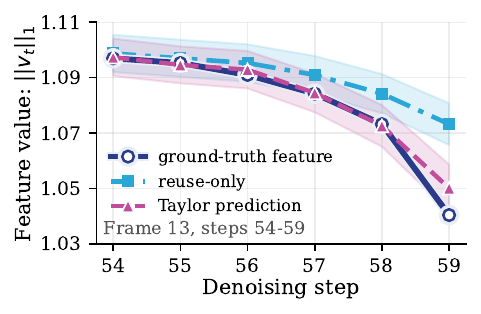}%
  }\hfill
  \subcaptionbox{Feature error\label{fig:binary-gap-error}}{%
    \includegraphics[width=0.498\linewidth]{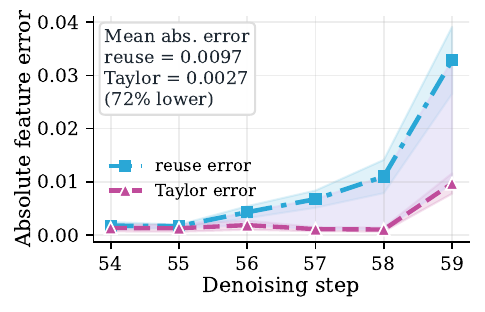}%
  }
  \caption{At Frame 13 over steps 54--59: (a)~Taylor prediction tracks the ground-truth feature more closely than reuse-only; (b)~Taylor prediction also yields lower feature error.}
  \Description{Two plots for SkyReels-V2 540P generation. The curves denote cross-prompt means and the shaded bands denote cross-prompt standard deviations. The left plot compares the feature value norm of the ground-truth feature, a reuse-only approximation, and Taylor prediction at Frame 13 over denoising steps 54 to 59, with Taylor prediction staying closer to the ground truth. The right plot compares mean absolute feature errors for reuse-only and Taylor prediction, showing lower error for Taylor prediction.}
  \label{fig:binary-gap}
\end{figure}

This suggests that some intermediate steps need not be treated as either perfect reuse or full recomputation. Instead, they can be approximated from cached history by extrapolating over nearby noise levels. As illustrated in Figure~\ref{fig:binary-gap}, on SkyReels-V2 at Frame 13 over steps 54--59, the cross-prompt mean trajectory of Taylor prediction stays closer to the ground-truth feature than reuse-only. The corresponding cross-prompt mean feature error is also consistently lower, reducing the mean absolute deviation by 72\% relative to reuse-only, while the shaded bands visualize the cross-prompt standard deviation under a single fixed seed over the 16-prompt diagnostic subset.

\begin{figure}[t]
  \centering
  \includegraphics[width=\linewidth]{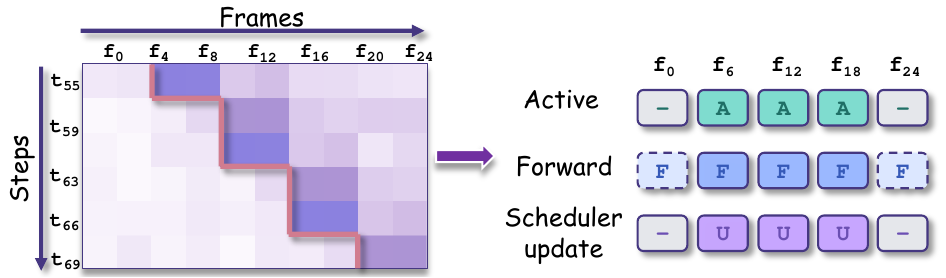}
  \caption{Conceptual step-matrix view of asynchronous autoregressive denoising. At each iteration, only part of the valid frame interval is active; dark cells mark frames whose scheduling state is still advancing.}
  \Description{An illustration of an asynchronous autoregressive denoising schedule, highlighting the active frame interval used by selective computation.}
  \label{fig:step-matrix}
\end{figure}

\textbf{Active frame interval.} We define the active frame interval directly from the scheduler-derived update mask. A frame is labeled active at iteration $k$ if and only if its scheduling state advances relative to iteration $k{-}1$ and it has not yet reached the terminal state. Figure~\ref{fig:step-matrix} visualizes this as a moving active suffix within the scheduler-valid interval. In the Forward row, solid boxes mark active frames, while dashed boxes mark forwarded but inactive ones. The model forward pass and per-frame state update are not always aligned under the asynchronous AR schedule, as some forwarded frames do not actually advance their denoising state, motivating selective computation to skip such redundant work.

\begin{figure*}[t]
  \centering
  \includegraphics[width=\textwidth]{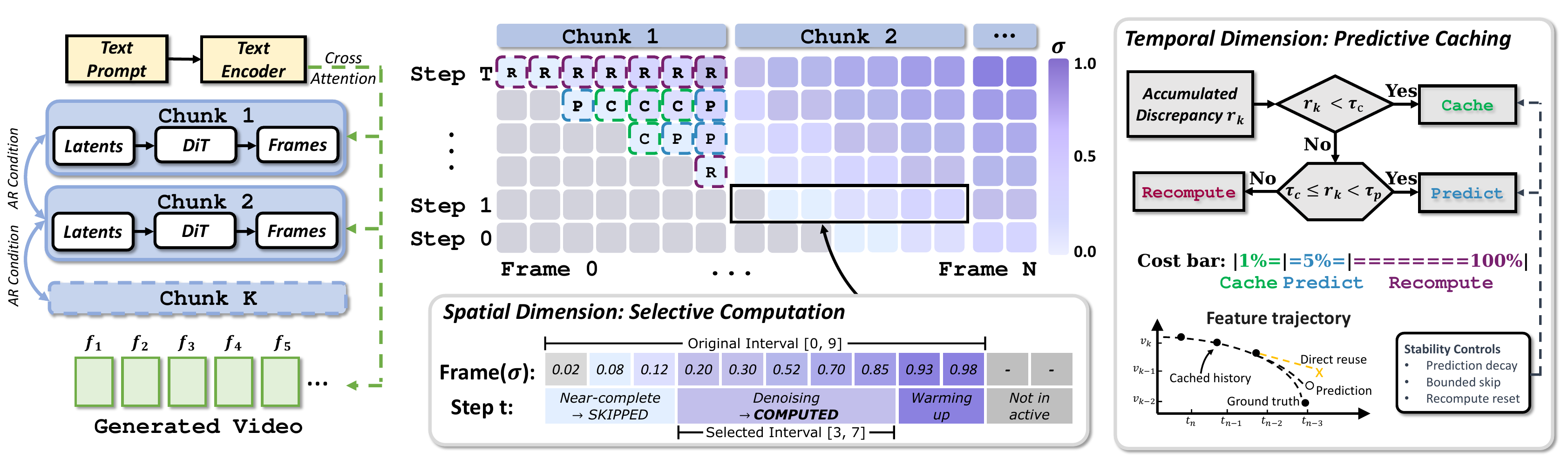}
  \caption{Overview of \method for accelerating autoregressive video diffusion by reducing redundant computation in both the spatial and temporal dimensions through selective computation and predictive extrapolation.}
  \Description{A framework figure for SCOPE. The left part shows autoregressive video generation across multiple chunks with denoising inside each chunk. The middle top shows a frame-by-step denoising matrix where different frames are at different stages. The middle bottom shows selective computation over the active frame interval. The right part shows predictive extrapolation along the temporal denoising trajectory using cached history.}
  \label{fig:method}
\end{figure*}

\subsection{Overview of \method}

Figure~\ref{fig:method} summarizes how \method accelerates autoregressive video diffusion through two complementary strategies. Selective computation removes redundant updates across frames, whereas predictive extrapolation reduces redundant computation across denoising.

Each chunk is denoised by a DiT conditioned on the text prompt and the decoded frames from the preceding chunk. At each denoising step, an accumulated discrepancy score $r_k$ is compared with thresholds $\tau_c$ and $\tau_p$ to select one of three modes along the denoising trajectory. In \predict{} mode, \method extrapolates cached velocity history via Taylor expansion, yielding a closer approximation to the ground truth trajectory than direct reuse. In parallel, selective computation leverages the asynchronous noise schedule across frames, so only frames that remain within the active denoising range are evaluated, while frames that are nearly complete or just beginning to denoise are skipped.

\subsection{Predictive Extension of Cache Reuse}

The temporal branch of \method extends existing cache-reuse acceleration. Traditional caching frameworks decide between direct reuse and full recomputation. \method introduces a third mode, \predict, which estimates the current velocity from cached history. Let $r_k$ denote the accumulated discrepancy score before the decision at step $k$, let $\tau_c^{(k)}$ and $\tau_p^{(k)}$ be the cache and prediction thresholds, and let $H_k$ be the maximum allowed non-recompute horizon. When cached output is available and prediction history is sufficient, the candidate mode is:
\begin{equation}
  m_k^{\mathrm{cand}} =
  \begin{cases}
    \cache, & r_k < \tau_c^{(k)} \text{ and cache available}, \\
    \predict, & \tau_c^{(k)} \le r_k < \tau_p^{(k)} \text{ and history available}, \\
    \recompute, & \text{otherwise}.
  \end{cases}
  \label{eq:trimodal-decision}
\end{equation}

This formulation fills the gap between zero-cost reuse and full-cost recomputation. The goal of \predict is not to replace recomputation globally, but to extend cache/reuse to cases where direct reuse is too coarse and full recomputation is unnecessarily expensive.

\textbf{Discrepancy estimation.}
We estimate an accumulated discrepancy score from lightweight features that can be computed before a full forward pass. Let $\phi_k$ be the host feature used for decision making at step $k$ and define the normalized discrepancy:
\begin{equation}
  d_k = \frac{\lVert \phi_k - \phi_{k-1} \rVert_1}{\lVert \phi_{k-1} \rVert_1 + \varepsilon}, 
\end{equation}
where $\varepsilon$ is a small positive constant for numerical stability. The implementation then updates an accumulated discrepancy score:
\begin{equation}
  \begin{aligned}
    r_k^- &= r_{k-1} + d_k, \\
    r_k &=
    \begin{cases}
      r_k^-, & m_k = \cache, \\
      \lambda r_k^-, & m_k = \predict, \\
      0, & m_k = \recompute, 
    \end{cases}
  \end{aligned}
  \label{eq:risk}
\end{equation}
where $\lambda \in (0,1]$ is the \texttt{predict\_decay}.

\textbf{Prediction branch.}
When a step enters \predict, \method forecasts the velocity from cached history in the noise-level coordinate. Let $\sigma^\star = g(t^\star)$ be the target noise level, let $(\sigma_k, v_k)$ be the most recent cached state, and define $\Delta \sigma_k = \sigma^\star - \sigma_k$, $\Delta \sigma_{k-1} = \sigma_k - \sigma_{k-1}$, and $\Delta \sigma_{k-2} = \sigma_{k-1} - \sigma_{k-2}$. The first-order predictor is:
\begin{equation}
  \hat{v}^{(1)}(\sigma^\star) = v_k + \frac{v_k - v_{k-1}}{\Delta \sigma_{k-1}} \Delta \sigma_k.
  \label{eq:first-order}
\end{equation}

With one more history point, the implementation approximates curvature by a second finite difference and obtains:
\begin{equation}
  \hat{v}^{(2)}(\sigma^\star) = \hat{v}^{(1)}(\sigma^\star) + \frac{1}{2}
  \frac{\frac{v_k - v_{k-1}}{\Delta \sigma_{k-1}} - \frac{v_{k-1} - v_{k-2}}{\Delta \sigma_{k-2}}}{(\Delta \sigma_{k-1} + \Delta \sigma_{k-2})/2}
  \Delta \sigma_k^2.
  \label{eq:second-order}
\end{equation}

In practice, the \predict branch uses a Taylor-based extrapolator with stability checks and conservative clipping.

% \textbf{Why prediction is a separate mode.}
% Forecasting is qualitatively different from direct reuse. Reuse assumes that the cached value remains valid. Prediction instead assumes that the trajectory is smooth and updates the cached state with a lightweight approximation. Treating prediction as its own mode allows the scheduler to operate on a richer cost-quality spectrum rather than a binary choice. The complete per-step algorithm consolidating both mechanisms is provided in the supplementary material.

\subsection{Selective Computation}

In asynchronous AR schedules, only a subset of frames remains actively denoising at each iteration. We identify this active subset directly from the scheduler step matrix. Let $u_{t,j}$ denote the scheduler progress index of latent slot $j$ at iteration $t$, let $N$ be the terminal progress index, let $F$ be the number of latent slots in the current scheduler segment, and let $B$ be the default scheduler window length. The update mask and the interval end $\bar{e}_t$ are:
\begin{equation}
  \begin{aligned}
    m_{t,j} &= \mathbb{1}[u_{t,j} \ne u_{t-1,j} \ \wedge \ u_{t,j} \ne N], \\
    \bar{e}_t &=
    \begin{cases}
      \min(B, F), & \{j : m_{t,j}=1\} = \emptyset, \\
      \max(\bar{e}_{t-1},\; 1 + \max\{j : m_{t,j}=1\}), & \text{otherwise},
    \end{cases}
  \end{aligned}
  \label{eq:activity-mask}
\end{equation}
where $\mathbb{1}[\cdot]$ denotes the indicator function, which equals 1 when the enclosed condition is true and 0 otherwise. $m_{t,j}$ is the scheduler-derived update mask indicating whether latent slot $j$ both advances at iteration $t$ and has not yet reached the terminal state. The resulting scheduler-facing valid interval is:
\begin{equation}
  V_t = [\max(\bar{e}_t - B, 0),\; \bar{e}_t).
  \label{eq:raw-interval}
\end{equation}

Relative to the baseline window length $B_t = \min(B, F)$ without selective computation, the per-step cost scales primarily with $|V_t|$, while the overhead of extracting the interval remains lower order. Because $\bar{e}_t$ is monotone, the compute interval remains a safe suffix of the scheduler-valid interval throughout the rollout. Selective computation therefore leaves the AR schedule itself unchanged, as it replaces the default full-length interval with a mask-derived compute interval. This mechanism is orthogonal to the predictive extension of cache reuse, so a step can be processed on a reduced interval. The detailed cost relation is provided in the supplementary material.

\begin{table*}[t]
  \caption{Main comparison on SkyReels-V2 DF-1.3B and MAGI-1 4.5B-distill, with all results measured on a single NVIDIA A800 80GB GPU. LPIPS, SSIM, and PSNR are computed against the Original output as reference. Best and second-best accelerated results are shown in bold and underlined, respectively. Original is excluded from ranking.}
  \label{tab:main-results}
  \centering
  \small
  \begin{tabular}{llcccccccc}
    \toprule
    \multirow{2}{*}{\textbf{\boldmath Model}} & \multirow{2}{*}{\textbf{\boldmath Method}} & \multicolumn{4}{c}{\textbf{\boldmath Efficiency}} & \multicolumn{4}{c}{\textbf{\boldmath Visual Quality}} \\
    \cmidrule(lr){3-6} \cmidrule(lr){7-10}
    & & \textbf{\boldmath Time (s)$\downarrow$} & \textbf{\boldmath Speedup$\uparrow$} & \textbf{\boldmath FLOPs (P)$\downarrow$} & \textbf{\boldmath Speedup$\uparrow$} & \textbf{\boldmath VBench$\uparrow$} & \textbf{\boldmath LPIPS$\downarrow$} & \textbf{\boldmath SSIM$\uparrow$} & \textbf{\boldmath PSNR$\uparrow$} \\
    \midrule
    \multirow{10}{*}{SkyReels-V2 DF-1.3B} & Original & 1452.41 & 1.00$\times$ & 245.01 & 1.00$\times$ & 81.51\% & 0.0000 & 1.0000 & $\infty$ \\
    \cmidrule(lr){2-10}
    & $\Delta$-DiT & 1223.01 & 1.19$\times$ & 201.60 & 1.22$\times$ & 74.57\% & 0.6052 & 0.5228 & 14.25 \\
    & TeaCache-slow & 697.60 & 2.08$\times$ & 109.44 & 2.24$\times$ & 79.83\% & 0.4436 & 0.5986 & 15.60 \\
    & TeaCache-fast & 559.76 & 2.59$\times$ & 84.68 & 2.89$\times$ & 77.49\% & 0.4786 & 0.5595 & 15.59 \\
    & TaylorSeer-slow & 941.41 & 1.54$\times$ & 130.76 & 1.87$\times$ & 75.54\% & 0.6838 & 0.4516 & 10.32 \\
    & TaylorSeer-fast & 799.55 & 1.82$\times$ & 99.05 & 2.47$\times$ & 75.50\% & 0.7017 & 0.4484 & 10.08 \\
    & FlowCache-slow & 429.50 & 3.38$\times$ & 54.62 & 4.49$\times$ & 80.04\% & 0.1978 & 0.6654 & 21.73 \\
    & FlowCache-fast & 364.72 & 3.98$\times$ & \textbf{42.56} & \textbf{5.76}$\times$ & 78.75\% & 0.2441 & 0.6058 & 18.92 \\
    \cmidrule(lr){2-10}
    & \method-slow & \underline{350.12} & \underline{4.15}$\times$ & 53.87 & 4.55$\times$ & \textbf{81.57\%} & \textbf{0.1676} & \textbf{0.8101} & \textbf{24.76} \\
    & \method-fast & \textbf{307.34} & \textbf{4.73}$\times$ & \underline{46.45} & \underline{5.27}$\times$ & \underline{81.47\%} & \underline{0.1900} & \underline{0.7309} & \underline{21.96} \\
    \midrule
    \multirow{10}{*}{MAGI-1 4.5B-distill} & Original & 791.08 & 1.00$\times$ & 81.09 & 1.00$\times$ & 76.65\% & 0.0000 & 1.0000 & $\infty$ \\
    \cmidrule(lr){2-10}
    & $\Delta$-DiT & 463.01 & 1.71$\times$ & 38.49 & 2.11$\times$ & 74.54\% & 0.3607 & 0.6473 & 17.50 \\
    & TeaCache-slow & 617.22 & 1.28$\times$ & 56.53 & 1.43$\times$ & \textbf{76.94\%} & 0.2686 & 0.6899 & 19.58 \\
    & TeaCache-fast & 525.34 & 1.51$\times$ & 43.55 & 1.86$\times$ & 76.20\% & 0.4274 & 0.5197 & 16.15 \\
    & TaylorSeer-slow & 486.79 & 1.63$\times$ & 39.14 & 2.07$\times$ & 76.05\% & 0.4935 & 0.3916 & 12.65 \\
    & TaylorSeer-fast & 460.62 & 1.72$\times$ & 35.54 & 2.28$\times$ & 75.29\% & 0.5005 & 0.3656 & 12.44 \\
    & FlowCache-slow & 535.89 & 1.48$\times$ & 41.62 & 1.95$\times$ & 76.09\% & 0.2619 & \underline{0.6960} & \underline{19.62} \\
    & FlowCache-fast & 451.08 & 1.75$\times$ & \underline{28.50} & \underline{2.85}$\times$ & 74.88\% & 0.4330 & 0.5142 & 15.99 \\
    \cmidrule(lr){2-10}
    & \method-slow & \underline{361.83} & \underline{2.19}$\times$ & 30.50 & 2.66$\times$ & \underline{76.46\%} & \textbf{0.2146} & \textbf{0.7844} & \textbf{23.56} \\
    & \method-fast & \textbf{310.16} & \textbf{2.55}$\times$ & \textbf{24.42} & \textbf{3.32}$\times$ & 76.32\% & \underline{0.2456} & 0.6499 & 17.06 \\
    \bottomrule
\end{tabular}
\end{table*}

\subsection{Error Analysis and Stability Controls}

\textbf{Error propagation analysis.}
To understand how prediction errors affect the output, we analyze how errors propagate through the denoising trajectory. When \method uses \predict mode, the predicted velocity $\hat{v}_k$ differs from the true velocity $v_k$, and this error compounds over multiple steps.

Let $e_k^v = \lVert \hat{v}_k - v_k \rVert_2$ denote the prediction error on velocity and let $e_k^x = \lVert \hat{x}_k - x_k \rVert_2$ denote the latent-state error before step $k$. From the Euler update in Eq.~\eqref{eq:euler-update}, the one-step latent drift satisfies:
\begin{equation}
  e_{k-1}^x \le e_k^x + |\Delta \sigma_k| e_k^v.
  \label{eq:latent-drift-onestep}
\end{equation}

This shows that each prediction error $e_k^v$ contributes to the latent error proportionally to the step size $|\Delta \sigma_k|$. After $h$ consecutive predicted steps, the cumulative error becomes:
\begin{equation}
  e_{k-h}^x \le e_k^x + \sum_{i=0}^{h-1} |\Delta \sigma_{k-i}| e_{k-i}^v.
  \label{eq:latent-drift-hstep}
\end{equation}

If the local first-order predictor obeys a Taylor remainder bound $e_k^v \le \frac{L_2}{2} |\Delta \sigma_k|^2$, then:

\begin{equation}
  e_{k-h}^x \le e_k^x + \frac{L_2}{2} \sum_{i=0}^{h-1} |\Delta \sigma_{k-i}|^3.
  \label{eq:drift-first-order}
\end{equation}

This bound reveals that the error grows cubically with step size, motivating three complementary stability controls.

\textbf{Stability controls.}
First, after each \predict step the discrepancy score is damped by a decay factor $r_k = \lambda r_k^-$ with $\lambda \in (0,1]$, steering the system toward a \recompute step before drift becomes excessive. Second, the maximum number of consecutive non-\recompute steps is capped at $H_k$, guaranteeing periodic updates regardless of the current discrepancy score. Third, every \recompute step resets the discrepancy score to $r_k = 0$ and refreshes the velocity cache, preventing error propagation across recompute boundaries. The full derivation is deferred to the supplementary material.

% Under consecutive predictive steps, Eq.~\eqref{eq:risk} implies:
% \begin{equation}
%   r_{k+h} = \lambda^h r_k + \sum_{i=1}^{h} \lambda^{h-i+1} d_{k+i},
%   \label{eq:risk-unroll}
% \end{equation}
% so the influence of past discrepancies on the accumulated risk decays geometrically rather than accumulating linearly. In practice, we additionally maintain an exponential moving average (EMA) of recent discrepancies, $s_k = \beta s_{k-1} + (1-\beta)d_k$, and use the guarded score $q_k = \max(r_k^-, s_k)$ for decisions near $\tau_c$ and $\tau_p$. This conservative override suppresses overly aggressive prediction after bursts of larger change, while the hard skip limit $H_k$ guarantees periodic recomputation even when the instantaneous discrepancy remains small. Together, decay, EMA guarding, and bounded skip length confine drift within each non-recompute segment while preserving the efficiency gains of \predict; the full derivation is deferred to the supplementary material.

\begin{figure*}[t]
  \centering
  \includegraphics[width=\textwidth,page=1]{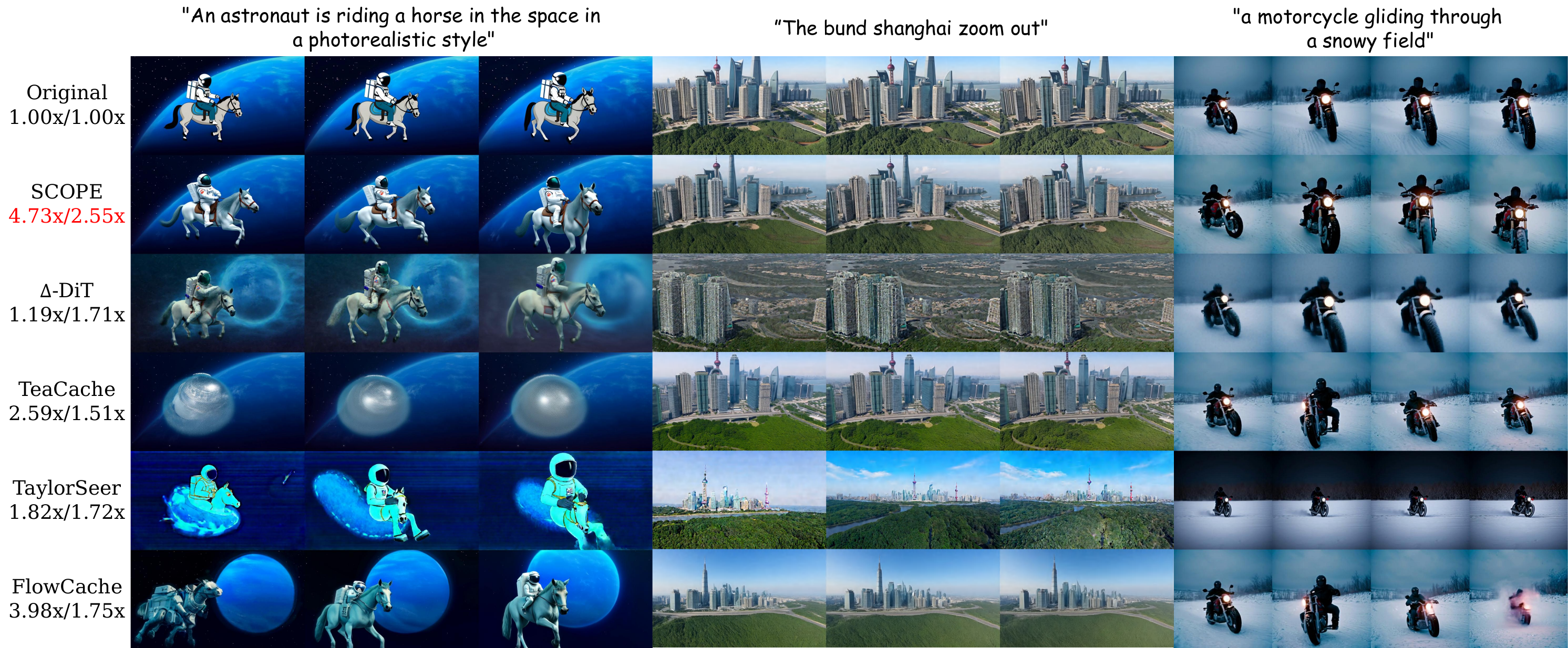}
  \caption{Qualitative comparison of all methods on representative prompts from the two models. The left two groups show results on SkyReels-V2, while the right group shows results on MAGI-1. Each method row is annotated with its end-to-end speedup on SkyReels-V2 and MAGI-1, respectively. Compared with other accelerated baselines, \method preserves object structure and visual fidelity more consistently while maintaining the strongest overall acceleration-quality tradeoff.}
  \Description{Representative qualitative comparison across all methods on the two target models, with each row annotated by its speedup on SkyReels-V2 and MAGI-1.}
  \label{fig:main_qualitative_20260321}
\end{figure*}

\begin{figure}[t]
  \centering
  \includegraphics[width=\linewidth]{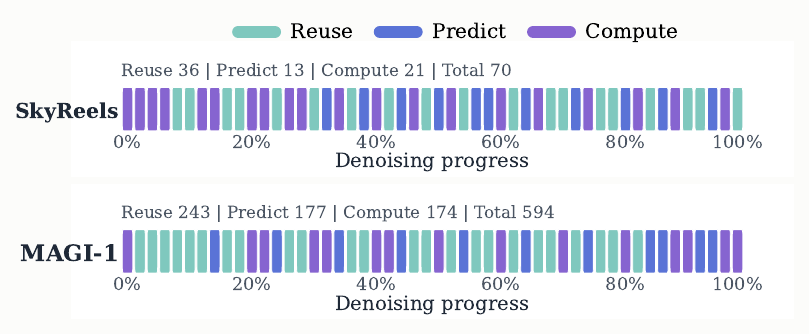}
  \caption{Tri-modal decision timeline comparison between SkyReels-V2 (top) and MAGI-1 (bottom). Each row shows the \cache, \predict, and \recompute decisions across denoising iterations.}
  \Description{Decision timeline visualization showing cache, predict, and recompute mode distributions across denoising iterations for both MAGI-1 and SkyReels-V2.}
  \label{fig:exp5_dual_model_decision_timeline_20260317}
\end{figure}

\section{Experiments}

This section evaluates \method on two representative AR diffusion models. We organize the experiments around the speed-quality tradeoff and the model-specific contribution of each technical line, reporting both quantitative comparisons and qualitative results.

\subsection{Models, Dataset, and Metrics}

We evaluate \method on two autoregressive video diffusion models. \textbf{SkyReels-V2 DF-1.3B}: we use the 540P diffusion-forcing model with the aligned benchmark configuration that generates 257-frame videos with a base window of 97 frames, 50 denoising steps, and overlap history 17. \textbf{MAGI-1 4.5B-distill}: we use the aligned distill runtime configuration with 240 frames at 480$\times$480 resolution and 64 denoising steps. KV offload remains disabled so that the Original and accelerated runs share the same basic generation setup.

The prompts are drawn from the VBench gallery~\cite{huang2024vbench}, spanning 16 evaluation dimensions. All accelerated methods match the Original pipeline exactly in frame count, resolution, denoising steps, seeds, and AR settings. A run is valid only after checking for failure modes such as black screens, frozen output, or severe artifacts.

\begin{table}[t]
  \centering
  \small
  \caption{Sensitivity analysis across threshold settings on both models. Each configuration is denoted as $(\tau_c, \tau_p)$ where $\tau_c$ is the cache threshold and $\tau_p$ is the prediction threshold.}
  \label{tab:exp3_threshold_longtable_20260317}
  \begin{tabular}{lccccc}
  \toprule
  \multirow{2}{*}{Config $(\tau_c, \tau_p)$} & \multicolumn{2}{c}{Efficiency} & \multicolumn{3}{c}{Quality} \\
  \cmidrule(lr){2-3} \cmidrule(lr){4-6}
  & Time (s)$\downarrow$ & Speedup$\uparrow$ & LPIPS$\downarrow$ & SSIM$\uparrow$ & PSNR$\uparrow$ \\
  \midrule
  \multicolumn{6}{c}{\textbf{SkyReels-V2}} \\
  \cmidrule(lr){1-6}
  $(0.12, 0.30)$ & 353.11 & 4.11$\times$ & 0.1723 & 0.7581 & 23.17 \\
  $(0.15, 0.36)$ & 326.21 & 4.45$\times$ & 0.1798 & 0.7460 & 22.74 \\
  $(0.18, 0.42)$ & 307.34 & 4.73$\times$ & 0.1900 & 0.7309 & 21.96 \\
  $(0.21, 0.48)$ & 281.56 & 5.16$\times$ & 0.2152 & 0.7018 & 20.69 \\
  $(0.25, 0.55)$ & 262.02 & 5.54$\times$ & 0.2399 & 0.6650 & 19.48 \\
  \midrule
  \multicolumn{6}{c}{\textbf{MAGI-1}} \\
  \cmidrule(lr){1-6}
  $(0.02, 0.05)$ & 362.34 & 2.18$\times$ & 0.2146 & 0.7844 & 23.56 \\
  $(0.03, 0.08)$ & 310.16 & 2.55$\times$ & 0.2456 & 0.6499 & 17.06 \\
  $(0.05, 0.12)$ & 271.46 & 2.91$\times$ & 0.2700 & 0.6174 & 16.98 \\
  $(0.08, 0.15)$ & 261.07 & 3.03$\times$ & 0.3077 & 0.5704 & 16.72 \\
  \bottomrule
  \end{tabular}
  \end{table}

We track three categories of metrics. For \textbf{efficiency}, we report end-to-end inference time and speedup relative to Original. For \textbf{quality}, we use VBench~\cite{huang2024vbench} total score and its 16 sub-dimensions, and we additionally record LPIPS~\cite{zhang2018unreasonable}, SSIM~\cite{wang2004image}, and PSNR as supplementary diagnostics. For \textbf{memory}, we report peak GPU memory during inference.

\subsection{Baselines}

Our main comparison includes the following training-free baselines. \textbf{Original} is the unaccelerated inference pipeline. \textbf{$\Delta$-DiT}~\cite{chen2025deltadit} performs block-level reuse adapted to the target AR pipelines by applying block-skip decisions per AR unit rather than globally. \textbf{TeaCache}~\cite{liu2025teacache} performs binary timestep-aware caching based on step similarity estimation. \textbf{TaylorSeer}~\cite{liu2025taylorseers} is a forecast-style accelerator. \textbf{FlowCache}~\cite{ma2026flowcache} applies chunk-aware adaptive caching for autoregressive video generation.   \textbf{\method} is our predictive caching method with selective computation. Detailed per-method hyperparameter settings are listed in the supplementary material.

\subsection{Implementation Details}

The default \method configuration uses a cache threshold $\tau_c$ and a prediction threshold $\tau_p$ to partition steps into \cache, \predict, and \recompute. On SkyReels-V2, the paper-core configuration uses $\tau_c=0.18$, $\tau_p=0.42$, second-order Taylor prediction, prediction decay 0.75, bounded skip length 5, and selective computation enabled. On MAGI-1, $\tau_c=0.03$, $\tau_p=0.08$, second-order Taylor prediction, prediction decay 0.75, and bounded skip length 5. These paper-core settings are chosen as practical operating points on the speed-quality frontier rather than to optimize a single metric. The sensitivity analysis in subsection~\ref{subsec:Sensitivity} shows the resulting tradeoff is smooth and the method is robust within a reasonable range. For each model, method, and configuration, we generate over 400 videos to ensure statistically meaningful comparisons. FlashAttention~\cite{dao2022flashattention} is enabled by default for all experiments. All experiments are conducted on a single NVIDIA A800 80GB GPU under the same inference setting. 

\begin{figure}[t]
  \centering
  \subcaptionbox{Time\label{fig:skip_sweep_time}}[0.49\columnwidth]{%
    \includegraphics[width=\linewidth]{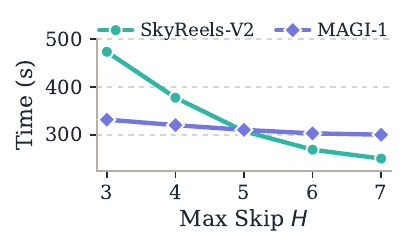}%
  }\hfill
  \subcaptionbox{1-LPIPS\label{fig:skip_sweep_1minus_lpips}}[0.49\columnwidth]{%
    \includegraphics[width=\linewidth]{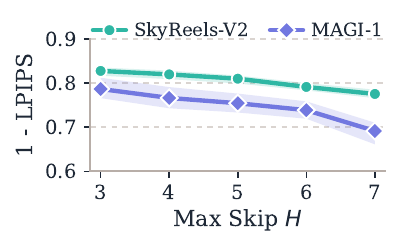}%
  }
  \subcaptionbox{SSIM\label{fig:skip_sweep_ssim}}[0.49\columnwidth]{%
    \includegraphics[width=\linewidth]{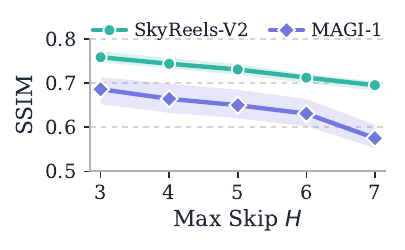}%
  }\hfill
  \subcaptionbox{PSNR\label{fig:skip_sweep_psnr}}[0.49\columnwidth]{%
    \includegraphics[width=\linewidth]{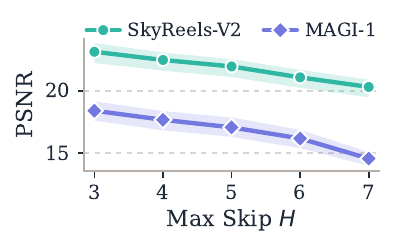}%
  }
  \caption{Bounded skip-length sensitivity analysis visualized across runtime and quality metrics for SkyReels-V2 and MAGI-1.}
  \Description{A single-column 2 by 2 figure for the bounded skip-length sensitivity analysis. The top-left panel shows Time, the top-right panel shows 1-LPIPS, the bottom-left panel shows SSIM, and the bottom-right panel shows PSNR for SkyReels-V2 and MAGI-1.}
  \label{fig:skip_sweep_metrics}
\end{figure}

\begin{figure}[t]
  \centering
  \begin{minipage}[t]{0.49\linewidth}
    \centering
    \includegraphics[width=\linewidth]{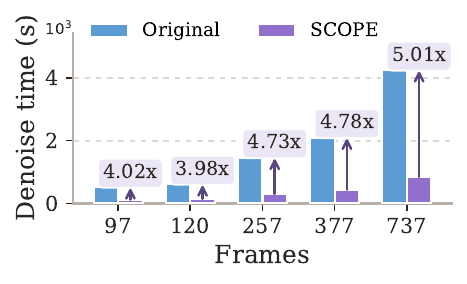}
  \end{minipage}
  \hfill
  \begin{minipage}[t]{0.49\linewidth}
    \centering
    \includegraphics[width=\linewidth]{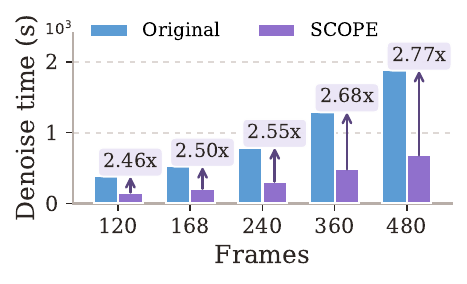}
  \end{minipage}
  \caption{Denoising time of Original versus \method at varying video lengths. Left: SkyReels-V2 DF-1.3B (97--737 frames). Right: MAGI-1 4.5B-distill (120--480 frames). Annotations indicate the speedup at each frame count.}
  \Description{Two side-by-side bar charts comparing denoising time of Original and SCOPE across different frame counts, with speedup annotations above each pair.}
  \label{fig:exp4_scalability_20260319}
\end{figure}

\subsection{Main Comparisons}

Table~\ref{tab:main-results} reports the main quantitative results on both target models, while Figure~\ref{fig:main_qualitative_20260321} shows representative qualitative comparisons. Across both SkyReels-V2 and MAGI-1, \method delivers the strongest overall speed-quality tradeoff among the accelerated baselines.

\textbf{Quantitative results.} On SkyReels-V2, \method-fast achieves $4.73\times$ speedup, significantly faster than FlowCache-fast. \method-slow reaches the highest VBench score among the accelerated methods at 81.57\%. This is numerically close to the Original score of 81.51\%; the 0.06 percentage-point difference lies within the natural variance of VBench, which aggregates 16 sub-dimensions where small perturbations in individual sub-dimension scores can cause minor fluctuations in either direction. We therefore interpret the two scores as effectively comparable. Methods that rely on wholesale prediction or simple block-level reuse show substantial quality degradation, confirming that a binary cache-or-recompute paradigm is insufficient for AR pipelines. On MAGI-1, \method-fast delivers $2.55\times$ speedup with 76.32\% VBench, while TeaCache-slow only achieves $1.28\times$ and FlowCache-fast reaches $1.75\times$ at a lower VBench of 74.88\%. The gap is especially pronounced in LPIPS and SSIM, where \method-slow outperforms the best competing method. The FLOPs reduction in Table~\ref{tab:main-results} provides a complementary view of efficiency. On SkyReels-V2, \method achieves $5.27\times$ FLOPs speedup versus $4.73\times$ time speedup, and on MAGI-1, it achieves $3.32\times$ versus $2.55\times$.

\textbf{Qualitative results.} The qualitative comparison in Figure~\ref{fig:main_qualitative_20260321} further reveals that \method preserves object structure, appearance fidelity, and scene consistency more reliably than aggressive reuse baselines. While TeaCache and TaylorSeer exhibit blurriness or structural distortion at higher speedups, \method maintains visual quality comparable to the original output. Additional visual comparisons are provided in the supplementary material.

% \textbf{FLOPs efficiency.} The FLOPs reduction in Table~\ref{tab:main-results} provides a complementary view of efficiency. On SkyReels-V2, \method achieves $5.27\times$ FLOPs speedup versus $4.73\times$ time speedup, and on MAGI-1, it achieves $3.32\times$ versus $2.55\times$. 
% This gap suggests that the remaining overhead comes from runtime factors such as memory transfers and scheduling, rather than model computation alone.

\begin{figure*}[t]
  \centering
  \begin{minipage}[c][0.22\textheight][t]{0.28\textwidth}
    \centering
    \includegraphics[width=\linewidth]{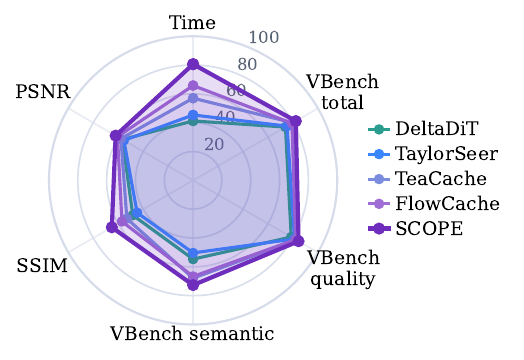}
    \caption{Radar comparison under time and quality metrics.}
    \label{fig:exp8_full_vbench_radar_skyreels_time_norm_quality_raw_20260320_180645}
    \Description{Radar chart comparing SkyReels-V2 quality and normalized runtime metrics across methods.}
  \end{minipage}
  \hfill
  \begin{minipage}[c][0.22\textheight][t]{0.34\textwidth}
    \centering
    \includegraphics[width=\linewidth,height=0.9\linewidth,keepaspectratio]{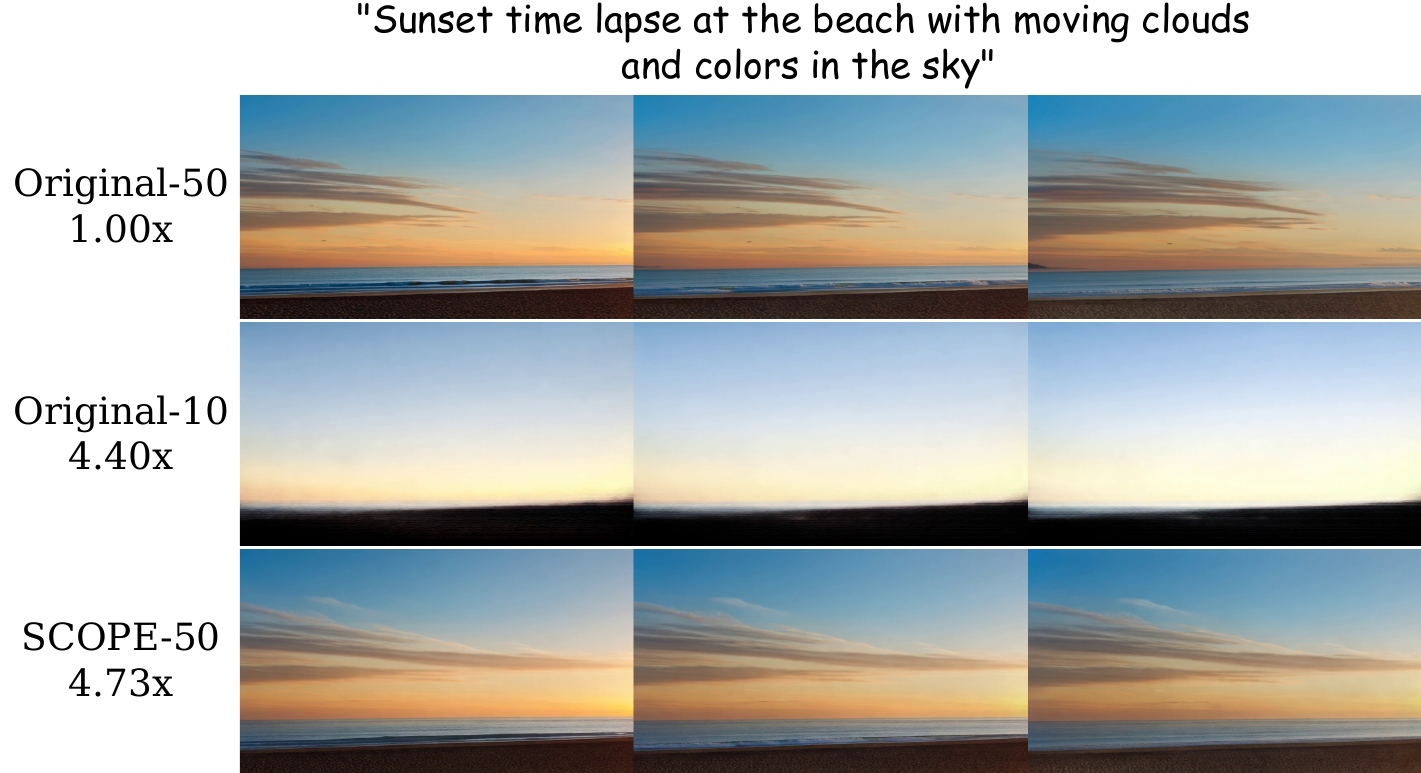}
    \caption{Few-step reduced-step Original comparison at matched runtime targets.}
    \label{fig:few-step}
    \Description{Comparison figure showing reduced-step Original behavior under few-step matched-time settings.}
  \end{minipage}
  \hfill
    \begin{minipage}[c][0.22\textheight][t]{0.36\textwidth}
    % \vspace{0pt}
    \centering
    \tiny
    \captionsetup*{type=table,skip=2pt}
    \caption{Same-time quality comparison between reduced-step Original and \method on both models.}
    \label{tab:exp10_same_time_original_20260319}
    % \resizebox{\linewidth}{!}{%
    \begin{tabular}{lccccc}
    \toprule
    \multirow{2}{*}{Method} & \multicolumn{2}{c}{Efficiency} & \multicolumn{3}{c}{Quality} \\
    \cmidrule(lr){2-3} \cmidrule(lr){4-6}
    & Time (s)$\downarrow$ & Speedup$\uparrow$ & LPIPS$\downarrow$ & SSIM$\uparrow$ & PSNR$\uparrow$ \\
    \midrule
    \multicolumn{6}{c}{\textbf{SkyReels-V2}} \\
    \cmidrule(lr){1-6}
    Original-50step & 1452.41 & 1.00$\times$ & 0.0000 & 1.0000 & $\infty$ \\
    Original-10step & 330.08 & 4.40$\times$ & 0.4523 & 0.5842 & 14.52 \\
    \method{}-50step & 307.34 & 4.73$\times$ & 0.1900 & 0.7309 & 21.96 \\
    \midrule
    \multicolumn{6}{c}{\textbf{MAGI-1}} \\
    \cmidrule(lr){1-6}
    Original-64step & 791.08 & 1.00$\times$ & 0.0000 & 1.0000 & $\infty$ \\
    Original-20step & 332.66 & 2.38$\times$ & 0.2912 & 0.6723 & 17.48 \\
    \method{}-64step & 310.16 & 2.55$\times$ & 0.2456 & 0.6499 & 17.06 \\
    \bottomrule
    \end{tabular}%
    % }
  \end{minipage}
  \end{figure*}

The radar chart for SkyReels-V2 in Figure~\ref{fig:exp8_full_vbench_radar_skyreels_time_norm_quality_raw_20260320_180645} further shows that \method occupies the strongest overall position across time-normalized efficiency and per-dimension quality metrics.
We also evaluated \method on the autoregressive image generation model Infinity~\cite{han2025infinity}, where it achieved a synchronized $1.56\times$ speedup with only small CLIPScore~\cite{radford2021learning} and ImageReward~\cite{xu2023imagereward} degradation, as detailed in the supplementary material.

\begin{figure}[t]
  \centering
  \includegraphics[width=\linewidth]{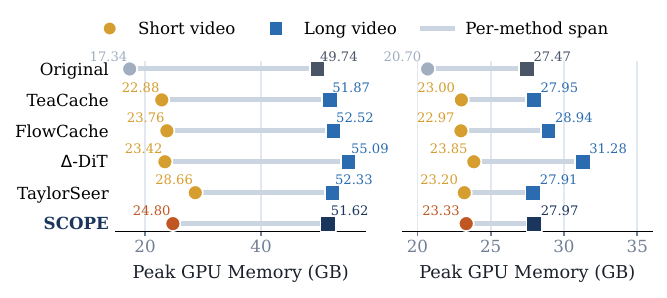}
  \caption{Peak GPU memory comparison under short and long video generation. Left: SkyReels-V2 at 257 and 480 frames. Right: MAGI-1 at 240 and 480 frames.}
  \label{fig:memory}
  \Description{A single-column memory comparison figure with two panels. The left panel shows short-video and long-video peak GPU memory for six methods on SkyReels-V2 at 257 and 480 frames, and the right panel shows the same comparison on MAGI-1 at 240 and 480 frames.}
  \end{figure}
  
\subsection{Ablation and Diagnostics}

\textbf{Selective computation.} Disabling selective computation reduces the speedup from $4.73\times$ to $3.72\times$ while improving LPIPS from 0.1900 to 0.1685, confirming that selective computation contributes a meaningful $1.27\times$ additional acceleration with a modest quality tradeoff. This shows that the spatial component provides an independent source of savings once the asynchronous frame activity is exposed. A stage-ratio study exploring different interval-ratio configurations is provided in the supplementary material.

\textbf{Predictor variants.} Comparing first-order Taylor, second-order Taylor, momentum, and auto-selection predictors reveals that the predictor choice has only a limited effect. On SkyReels-V2, all variants achieve $4.73$--$4.74\times$ speedup with LPIPS within 0.001 of each other. On MAGI-1, Taylor-2 provides a marginal improvement over Taylor-1. These results indicate that the main gain does not come from a fragile predictor choice. Instead, the contribution of the temporal branch is to turn prediction into a controlled intermediate mode inside the tri-modal scheduler, for which several simple extrapolators are already sufficient. The full predictor comparison table is provided in the supplementary material.

\textbf{Decision timeline.} Figure~\ref{fig:exp5_dual_model_decision_timeline_20260317} visualizes the tri-modal decision distribution over denoising iterations for MAGI-1 and SkyReels-V2. Total 70 and Total 594 denote the total number of SCOPE denoising decisions over the full generation process. MAGI-1 follows a chunk-synchronized schedule, with \cache concentrated in early low-discrepancy steps, \predict in the middle, and \recompute near chunk transitions. SkyReels-V2 instead exhibits an asynchronous staircase pattern, which creates more \predict opportunities in the early and middle stages while requiring more frequent \recompute near convergence. Despite these differences, the scheduler identifies intermediate steps suitable for \predict on both architectures. The larger share of skippable steps in SkyReels-V2 helps explain its higher time speedup under the same bounded skip horizon.

\subsection{Sensitivity Analysis}
\label{subsec:Sensitivity}

\textbf{Threshold sensitivity.} Table~\ref{tab:exp3_threshold_longtable_20260317} presents a sensitivity analysis across different cache thresholds $\tau_c$ and prediction thresholds $\tau_p$ for both models. On SkyReels-V2, increasing $(\tau_c, \tau_p)$ from (0.12, 0.30) to (0.25, 0.55) smoothly trades quality for speed, ranging from $4.11\times$ at 0.1723 LPIPS to $5.54\times$ at 0.2399 LPIPS. On MAGI-1, the tradeoff is steeper: the most aggressive setting (0.08, 0.15) reaches $3.03\times$ but with 0.3077 LPIPS. These results show that \method is robust to threshold selection within a reasonable range, with the paper-core settings representing a practical sweet spot.

\textbf{Bounded skip length.} We analyze the sensitivity of \method to the bounded skip length $H$ in Fig.~\ref{fig:skip_sweep_metrics}. Increasing $H$ consistently reduces runtime but also lowers $1-\mathrm{LPIPS}$, SSIM, and PSNR, making $H$ the main control knob for the efficiency-quality trade-off. The sensitivity is model dependent: MAGI-1 degrades more sharply as $H$ increases, whereas SkyReels-V2 remains more stable over moderate skip lengths. This suggests that $H$ should be tuned separately for each backbone. Exact numerical results are provided in the supplementary material.

\subsection{Scalability Analysis}

\textbf{Video length scaling.} Figure~\ref{fig:exp4_scalability_20260319} compares the denoising time of Original and \method across different video lengths on both target models. On SkyReels-V2, the speedup grows steadily from $4.02\times$ at 97 frames to $5.01\times$ at 737 frames. On MAGI-1, a similar trend holds: the speedup rises from $2.46\times$ at 120 frames to $2.77\times$ at 480 frames. In both cases, longer videos yield higher acceleration because additional AR chunks introduce more denoising iterations where the predictive caching policy can skip redundant computation. This confirms that \method scales favorably with video length rather than saturating at a fixed speedup ratio.

\textbf{Peak GPU memory.} Figure~\ref{fig:memory} reports peak GPU memory for short and long video generation, where short and long correspond to 257 and 480 frames on SkyReels-V2 and 240 and 480 frames on MAGI-1, respectively. On short videos, all caching-based methods incurred comparable overhead relative to Original. \method used 24.80\,GB versus 17.34\,GB on SkyReels-V2 and 23.33\,GB versus 20.70\,GB on MAGI-1, in line with TeaCache and FlowCache. On long videos, the memory gap narrowed as model activations dominated the budget, with all methods clustering between 49--55\,GB on SkyReels-V2 and 27--31\,GB on MAGI-1. $\Delta$-DiT showed the highest overhead on MAGI-1 due to its block-level reuse buffers. Overall, \method achieved the strongest speed-quality tradeoff without introducing disproportionate memory cost.

\subsection{Same-Time Original Step Reduction}

Table~\ref{tab:exp10_same_time_original_20260319} compares \method against reduced-step Original baselines matched to nearby runtime targets. For SkyReels-V2, the nearest reduced-step baseline uses 10 denoising steps, and for MAGI-1 it uses 20 steps. The results show that on SkyReels-V2, reduced-step Original is both slower and substantially lower in quality than the matched \method configuration. On MAGI-1, the reduced-step baseline is still slower and exhibits worse LPIPS. This gap arises because naively reducing denoising steps removes entire trajectory segments, breaking the smooth evolution that the ODE integrator relies on, whereas \method preserves the full trajectory and only approximates intermediate states via local extrapolation. Figure~\ref{fig:few-step} visualizes this comparison.

\section{Conclusion}

We presented \method, a training-free acceleration framework for autoregressive video diffusion that addressed temporal redundancy across denoising steps and spatial redundancy within the current valid interval. \method combines a tri-modal scheduler over \cache, \predict, and \recompute with selective computation that restricts execution to the active frame interval. Across MAGI-1 and SkyReels-V2, \method achieved up to $4.73\times$ speedup while preserving near-lossless VBench quality, outperforming all training-free baselines in the overall speed-quality tradeoff. A current limitation is that the operating point still requires per-model threshold calibration, which reduces direct transferability across architectures. Future work will explore learned predictor selection and integration with orthogonal attention-level acceleration to further improve generality.

%% Acknowledgments - comment out during anonymous submission
% \begin{acks}
% Acknowledge funding support and people who helped with your research.
% \end{acks}

%% References
\bibliographystyle{ACM-Reference-Format}
\bibliography{sample-base}
\clearpage
\appendix
\section{Overview}

This appendix supplements the main submission with the following material: (a)~full theoretical derivations of prediction error bounds, latent drift, and selective-computation cost in Section~\ref{sec:theory}; (b)~a consolidated algorithm listing for the \method tri-modal decision loop in Section~\ref{sec:algorithm}; (c)~evaluation metric definitions in Section~\ref{sec:metrics}; (d)~supplementary tables for baseline configurations, the stage-ratio study, predictor comparison, and bounded skip-length sensitivity analysis in Section~\ref{sec:suppl-notes}; (e)~an additional evaluation on autoregressive image generation with Infinity in Section~\ref{sec:ar-image}; and (f)~additional visual comparisons with representative frames across both target models in Section~\ref{sec:more-visual}.

\section{Theoretical Analysis}
\label{sec:theory}

\subsection{Local Error of Noise-Level Taylor Prediction}

Let $\sigma = g(t)$ be the scheduler-specific noise-level coordinate and let $\delta_k = \sigma^\star - \sigma_k$ be the target step gap. If $v$ is twice continuously differentiable on the interval between $\sigma_k$ and $\sigma^\star$, Taylor's theorem gives:
\begin{equation}
  v(\sigma^\star) = v(\sigma_k) + \frac{dv}{d\sigma}\bigg|_k \delta_k + \frac{1}{2} \frac{d^2 v}{d\sigma^2}\bigg|_{\xi_k} \delta_k^2,
\end{equation}
for some $\xi_k$ between $\sigma_k$ and $\sigma^\star$. Therefore, if $\left\lVert \frac{d^2 v}{d\sigma^2} \right\rVert_2 \le L_2$ locally, the first-order predictor defined in the main paper satisfies:
\begin{equation}
  \left\lVert v(\sigma^\star) - \hat{v}^{(1)}(\sigma^\star) \right\rVert_2 \le \frac{L_2}{2} |\delta_k|^2.
\end{equation}

If $v$ is three times continuously differentiable and the second-order correction is numerically stable, the second-order predictor obeys the tighter bound:
\begin{equation}
  \left\lVert v(\sigma^\star) - \hat{v}^{(2)}(\sigma^\star) \right\rVert_2 \le \frac{L_3}{6} |\delta_k|^3.
\end{equation}

These relations formalize why prediction is reliable only when the denoising trajectory is locally smooth and the scheduler gap is modest.

\subsection{Accumulated Latent Drift Under Consecutive Prediction}

The main paper derives the one-step latent drift, its $h$-step unrolling, and the first-order Taylor remainder bound. Here we extend the analysis to the second-order case. Substituting the second-order bound established above into the $h$-step recurrence yields:
\begin{equation}
  e_{k-h}^x \le e_k^x + \frac{L_3}{6} \sum_{i=0}^{h-1} |\Delta \sigma_{k-i}|^4,
\end{equation}
which provides a quartic rather than cubic dependence on step size, explaining the improved accuracy of second-order Taylor prediction observed experimentally.

% \subsection{Effect of Stability Controls}
%
% Let $r_k^- = r_{k-1} + \hat{e}_k$ denote the pre-decision accumulated risk and let $\lambda \in (0,1]$ denote the code-level \texttt{predict\_decay}. Recalling the risk-state update from the main paper, after each decision the accumulated risk evolves as:
% \begin{equation}
%   r_k =
%   \begin{cases}
%     r_k^-, & m_k = \cache, \\
%     \lambda r_k^-, & m_k = \predict, \\
%     0, & m_k = \recompute.
%   \end{cases}
% \end{equation}
%
% After $h$ consecutive predictive steps:
% \begin{equation}
%   r_{k+h} = \lambda^h r_k + \sum_{i=1}^{h} \lambda^{h-i+1}\hat{e}_{k+i}.
% \end{equation}
%
% Thus smaller $\lambda$ shortens the memory of past discrepancy. It adds a conservative override near threshold boundaries. Together with the hard skip limit, this yields the guarded scheduler used in the main paper.
%
% % The EMA guard is defined as:
% % \begin{equation}
% %   s_k = \beta s_{k-1} + (1-\beta)\hat{e}_k,
% %   \qquad
% %   q_k = \max(r_k^-, s_k).
% % \end{equation}

\subsection{Selective-Computation Cost Analysis}

The main paper defines the activity mask and interval extraction used in the paper-core setting. Here we provide the detailed cost relation deferred from the main text. Let $c_t$ denote the average per-slot cost of one denoising forward pass, $c_{\mathrm{ovh}}$ the interval-extraction overhead, $B_t = \min(B, F)$ the default window length, and $|V_t|$ the length of the selectively computed interval. The no-selective baseline costs:
\begin{equation}
  C_t^{\mathrm{full}} = c_t B_t,
\end{equation}
whereas selective computation satisfies:
\begin{equation}
  C_t^{\mathrm{sel}} \le c_t |V_t| + c_{\mathrm{ovh}}.
\end{equation}

Therefore, the cost ratio is bounded by:
\begin{equation}
  \frac{C_t^{\mathrm{sel}}}{C_t^{\mathrm{full}}} \le \frac{|V_t|}{B_t} + \frac{c_{\mathrm{ovh}}}{c_t B_t}.
\end{equation}

The speedup is largest when the active interval is much shorter than the scheduler-valid interval and the overhead remains negligible.

\section{Algorithm Implementation}
\label{sec:algorithm}

Algorithm~\ref{alg:scope} consolidates the \method per-step logic into a single listing organized around the two complementary acceleration mechanisms. The discrepancy metric $h(\cdot)$ is the calibration function that maps the lightweight feature difference to the discrepancy score used by the scheduler. The \texttt{TaylorPredict} subroutine implements the first- or second-order extrapolation defined in Eqs.~(6)--(7) of the main paper.

\begin{algorithm}[t]
\caption{\method Per-Step Decision and Execution}
\label{alg:scope}
\begin{algorithmic}[1]
\renewcommand{\algorithmicrequire}{\textbf{Input:}}
\renewcommand{\algorithmicensure}{\textbf{Output:}}
\Require Noisy latent $x_K$; velocity model $v_\theta$; denoising schedule $\{\sigma_k\}_{k=K}^{0}$; scheduler step matrix $\{u_{t,j}\}$; terminal index $N$; feature extractor $\phi$; thresholds $\tau_c, \tau_p$; decay $\lambda$; max skip $H$; window size $B$; total slots $F$
\Ensure Denoised latent $x_0$
\State $r \leftarrow 0$,\; $n_{\mathrm{skip}} \leftarrow 0$,\; history $\leftarrow \emptyset$
\For{$k = K$ \textbf{to} $1$}
  \Statex \hspace{\algorithmicindent}\Comment{\textbf{Selective Computation}}
  \State $m_{t,j} \leftarrow \mathbb{1}[u_{t,j} \ne u_{t-1,j} \;\wedge\; u_{t,j} \ne N]$ for each slot $j$
  \State $\bar{e}_t \leftarrow \max\bigl(\bar{e}_{t-1},\; 1 + \max\{j : m_{t,j}=1\}\bigr)$
  \State $V_t \leftarrow [\max(\bar{e}_t - B,\, 0),\; \bar{e}_t)$
  \Statex \hspace{\algorithmicindent}\Comment{\textbf{Predictive Caching}}
  \State $\hat{e}_k \leftarrow h\bigl(\lVert \phi_k - \phi_{k-1} \rVert_1 / (\lVert \phi_{k-1} \rVert_1 + \varepsilon)\bigr)$;\enspace $r \leftarrow r + \hat{e}_k$
  \If{$n_{\mathrm{skip}} \ge H$} \enspace $m_k \leftarrow \textsc{Recompute}$ \Comment{skip horizon}
  \ElsIf{$r < \tau_c$} \enspace $m_k \leftarrow \textsc{Cache}$
  \ElsIf{$r < \tau_p$} \enspace $m_k \leftarrow \textsc{Predict}$
  \Else \enspace $m_k \leftarrow \textsc{Recompute}$
  \EndIf
  \If{$m_k = \textsc{Cache}$}
    \State $v_k \leftarrow v_{k-1}$;\enspace $n_{\mathrm{skip}} \mathrel{+}= 1$
  \ElsIf{$m_k = \textsc{Predict}$}
    \State $v_k \leftarrow \text{TaylorPredict}(\text{history},\, \sigma^\star\!-\!\sigma_k)$;
    \State  $r \leftarrow \lambda r$;\enspace $n_{\mathrm{skip}} \mathrel{+}= 1$
  \Else
    \State $v_k \leftarrow v_{\theta}(x_k, \sigma_k)$ \textbf{on} $V_t$;\enspace $r \leftarrow 0$;\enspace $n_{\mathrm{skip}} \leftarrow 0$
    \State history $\leftarrow$ history $\cup\,(\sigma_k, v_k)$
  \EndIf
  \State $x_{k-1} \leftarrow x_k + v_k \cdot (\sigma_{k-1} - \sigma_k)$
\EndFor
\State \Return $x_0$
\end{algorithmic}
\end{algorithm}

\section{Evaluation Metrics}
\label{sec:metrics}

For each generated video $V_i = \{x_{i,t}\}_{t=1}^{T_i}$ and its matched reference video $V_i^\star = \{x_{i,t}^\star\}_{t=1}^{T_i}$, we compute LPIPS, SSIM, and PSNR frame by frame. In our experiments, the reference video is the Original output generated with the same prompt, seed, frame count, and resolution. For any frame-wise metric $M \in \{\mathrm{LPIPS}, \mathrm{SSIM}, \mathrm{PSNR}\}$, the video-level score is:
\begin{equation}
  M_i = \frac{1}{T_i} \sum_{t=1}^{T_i} M(x_{i,t}, x_{i,t}^\star),
\end{equation}
and the dataset-level score reported in the paper is the mean over the $N$ evaluated videos,
\begin{equation}
  \bar{M} = \frac{1}{N} \sum_{i=1}^{N} M_i.
\end{equation}

LPIPS is computed from deep features as:
\begin{equation}
  \mathrm{LPIPS}(x, y) = \sum_{l} \frac{1}{H_l W_l}
  \left\lVert
    w_l \odot \left( \hat{\phi}_l(x) - \hat{\phi}_l(y) \right)
  \right\rVert_2^2,
\end{equation}
where $\hat{\phi}_l(\cdot)$ denotes the channel-wise normalized feature at layer $l$, $w_l$ is the learned calibration weight, and $H_l \times W_l$ is the spatial resolution of that feature map.

SSIM is defined as:
\begin{equation}
  \mathrm{SSIM}(x, y) =
  \frac{(2\mu_x \mu_y + c_1)(2\sigma_{xy} + c_2)}
       {(\mu_x^2 + \mu_y^2 + c_1)(\sigma_x^2 + \sigma_y^2 + c_2)},
\end{equation}
where $\mu_x$ and $\mu_y$ are local means, $\sigma_x^2$ and $\sigma_y^2$ are local variances, and $\sigma_{xy}$ is the local covariance.

PSNR is defined from the mean squared error as:
\begin{equation}
  \mathrm{MSE}(x, y) = \frac{1}{|\Omega|} \sum_{p \in \Omega} (x_p - y_p)^2,
\end{equation}
\begin{equation}
  \mathrm{PSNR}(x, y) = 10 \log_{10} \left( \frac{L^2}{\mathrm{MSE}(x, y)} \right),
\end{equation}
where $\Omega$ is the set of pixel locations and $L$ is the maximum pixel value.

\section{Supplementary Ablation Tables}
\label{sec:suppl-notes}

This section provides the detailed ablation tables referenced in the main paper: baseline configurations, the stage-ratio study for selective computation, the predictor variant comparison, and the bounded skip-length sensitivity analysis with exact numerical values.

\subsection{Baseline Configurations}

Table~\ref{tab:baseline-config} lists the key hyperparameters used for each baseline method on both target models. All baselines are tuned to produce two representative operating points (slow/conservative and fast/aggressive) where applicable.

\begin{table}[t]
\centering
\small
\caption{Baseline configurations on both target models.}
\label{tab:baseline-config}
\begin{tabular}{llll}
\toprule
Method & Model & Key Parameter & Value \\
\midrule
\multirow{2}{*}{\deltadit} & MAGI-1 & cache\_interval & 8 \\
& SkyReels-V2 & cache\_interval & 3 \\
\midrule
\multirow{2}{*}{TeaCache-slow} & MAGI-1 & rel\_l1\_thresh & 0.008 \\
& SkyReels-V2 & teacache\_thresh & 0.10 \\
\multirow{2}{*}{TeaCache-fast} & MAGI-1 & rel\_l1\_thresh & 0.015 \\
& SkyReels-V2 & teacache\_thresh & 0.15 \\
\midrule
\multirow{2}{*}{TaylorSeer-slow} & MAGI-1 & max\_forecast\_steps & 3 \\
& SkyReels-V2 & max\_forecast\_steps & 2 \\
\multirow{2}{*}{TaylorSeer-fast} & MAGI-1 & max\_forecast\_steps & 4 \\
& SkyReels-V2 & max\_forecast\_steps & 3 \\
\midrule
\multirow{2}{*}{FlowCache-slow} & MAGI-1 & rel\_l1\_thresh & 0.008 \\
& SkyReels-V2 & teacache\_thresh & 0.10 \\
\multirow{2}{*}{FlowCache-fast} & MAGI-1 & rel\_l1\_thresh & 0.015 \\
& SkyReels-V2 & teacache\_thresh & 0.15 \\
\bottomrule
\end{tabular}
\end{table}

\subsection{Stage-Ratio Study for Selective Computation}

Table~\ref{tab:stage-ratio-app} studies whether selective computation should be tuned differently across denoising stages in SkyReels-V2. Here M and L denote the interval ratios used in the middle and late denoising stages, defined by the raw scheduler timestep $t$ as $200 < t < 800$ and $t \le 200$, respectively; the early stage ($t \ge 800$) keeps the default full interval. More conservative interval ratios consistently slow inference and slightly hurt quality, while the default full-ratio setting gives the best overall operating point among the tested stage-aware variants.

\begin{table}[t]
\centering
\small
\caption{Stage-ratio study for selective computation on SkyReels-V2.}
\label{tab:stage-ratio-app}
\begin{tabular}{lccccc}
\toprule
\multirow{2}{*}{Config} & \multicolumn{2}{c}{Efficiency} & \multicolumn{3}{c}{Quality} \\
\cmidrule(lr){2-3} \cmidrule(lr){4-6}
& Time (s)$\downarrow$ & Speedup$\uparrow$ & LPIPS$\downarrow$ & SSIM$\uparrow$ & PSNR$\uparrow$ \\
\midrule
No selective & 390.12 & 3.72$\times$ & 0.1685 & 0.8053 & 24.42 \\
M\,0.55\,/\,L\,0.35 & 343.27 & 4.23$\times$ & 0.2198 & 0.6878 & 21.42 \\
M\,1.00\,/\,L\,0.75 & 318.48 & 4.56$\times$ & 0.1901 & 0.7302 & 22.03 \\
M\,1.00\,/\,L\,1.00 & 307.34 & 4.73$\times$ & 0.1900 & 0.7309 & 21.96 \\
\bottomrule
\end{tabular}
\end{table}

\subsection{Predictor Comparison}

Table~\ref{tab:predictor-app} compares predictor variants used in the \predict mode of \method. First-order, second-order, and momentum-style predictors land at similar operating points, showing that the scheduler design matters more than the exact extrapolator.

\begin{table}[t]
\centering
\small
\caption{Predictor comparison across both target models.}
\label{tab:predictor-app}
\begin{tabular}{lccccc}
\toprule
\multirow{2}{*}{Predictor} & \multicolumn{2}{c}{Efficiency} & \multicolumn{3}{c}{Quality} \\
\cmidrule(lr){2-3} \cmidrule(lr){4-6}
& Time (s)$\downarrow$ & Speedup$\uparrow$ & LPIPS$\downarrow$ & SSIM$\uparrow$ & PSNR$\uparrow$ \\
\midrule
\multicolumn{6}{c}{\textbf{SkyReels-V2}} \\
\cmidrule(lr){1-6}
auto & 329.67 & 4.41$\times$ & 0.1900 & 0.7313 & 21.97 \\
momentum & 306.70 & 4.74$\times$ & 0.1898 & 0.7317 & 21.98 \\
taylor1 & 306.41 & 4.74$\times$ & 0.1902 & 0.7308 & 21.97 \\
taylor2 & 307.34 & 4.73$\times$ & 0.1900 & 0.7309 & 21.96 \\
\midrule
\multicolumn{6}{c}{\textbf{MAGI-1}} \\
\cmidrule(lr){1-6}
taylor1 & 310.00 & 2.55$\times$ & 0.2460 & 0.6491 & 17.04 \\
taylor2 & 310.16 & 2.55$\times$ & 0.2456 & 0.6499 & 17.06 \\
\bottomrule
\end{tabular}
\end{table}

\subsection{Maximum Skip-Step Sweep}

Table~\ref{tab:max-skip-app} studies the effect of the bounded skip horizon. The runtime-quality trade-off is monotonic on both models, with increasingly aggressive skip limits providing diminishing speed gains at a noticeable quality cost.

\begin{table}[t]
\centering
\small
\caption{Effect of \texttt{max\_skip\_steps} on runtime and quality for \method.}
\label{tab:max-skip-app}
\begin{tabular}{lccccc}
\toprule
\multirow{2}{*}{Max Skip} & \multicolumn{2}{c}{Efficiency} & \multicolumn{3}{c}{Quality} \\
\cmidrule(lr){2-3} \cmidrule(lr){4-6}
& Time (s)$\downarrow$ & Speedup$\uparrow$ & LPIPS$\downarrow$ & SSIM$\uparrow$ & PSNR$\uparrow$ \\
\midrule
\multicolumn{6}{c}{\textbf{SkyReels-V2}} \\
\cmidrule(lr){1-6}
3 & 473.34 & 3.07$\times$ & 0.1724 & 0.7586 & 23.14 \\
4 & 377.42 & 3.85$\times$ & 0.1803 & 0.7438 & 22.48 \\
5 & 307.34 & 4.73$\times$ & 0.1900 & 0.7309 & 21.96 \\
6 & 268.76 & 5.40$\times$ & 0.2087 & 0.7124 & 21.08 \\
7 & 250.18 & 5.81$\times$ & 0.2246 & 0.6953 & 20.31 \\
\midrule
\multicolumn{6}{c}{\textbf{MAGI-1}} \\
\cmidrule(lr){1-6}
3 & 331.57 & 2.39$\times$ & 0.2133 & 0.6862 & 18.40 \\
4 & 320.08 & 2.47$\times$ & 0.2336 & 0.6644 & 17.67 \\
5 & 310.16 & 2.55$\times$ & 0.2456 & 0.6499 & 17.06 \\
6 & 302.84 & 2.61$\times$ & 0.2613 & 0.6309 & 16.17 \\
7 & 300.14 & 2.64$\times$ & 0.3084 & 0.5753 & 14.55 \\
\bottomrule
\end{tabular}
\end{table}

\section{Acceleration on Autoregressive Image Generation}
\label{sec:ar-image}

We also evaluated \method on Infinity~\cite{han2025infinity}, a bitwise autoregressive image generation model, using synchronized timing for fair comparison. As reported in Table~\ref{tab:infinity-app}, the quality-first setting retains a measurable $1.56\times$ speedup relative to the matched Original baseline, with only small degradations in CLIPScore~\cite{radford2021learning} and ImageReward~\cite{xu2023imagereward}. This suggests that \method remains stable under a broader evaluation setup, while also indicating that the quality-first setting leaves limited headroom for further acceleration.

\begin{figure}[t]
  \centering
  \includegraphics[width=\columnwidth,page=1]{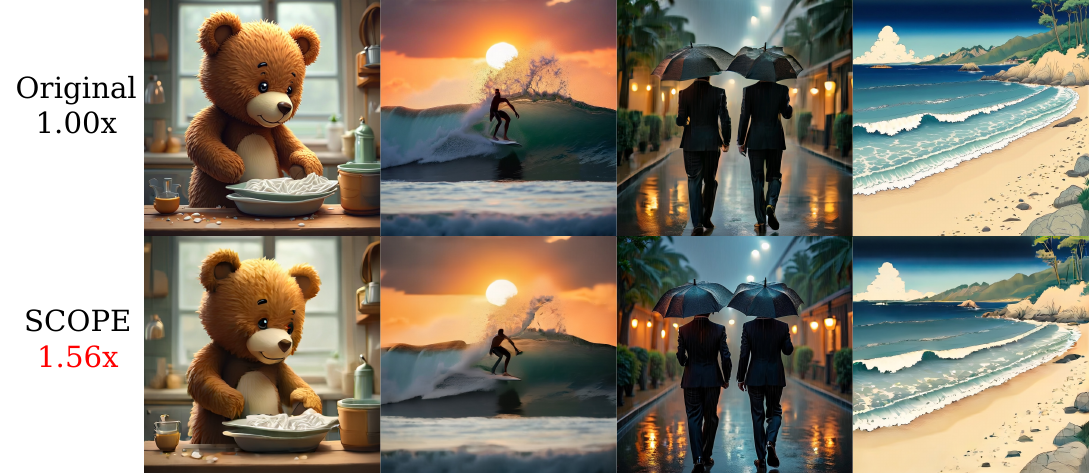}
  \caption{Visual comparison on Infinity under the quality-first setting. Zoom in for best viewing.}
  \Description{Single-column qualitative comparison on Infinity between the Original baseline and SCOPE.}
  \label{fig:infinity-image}
\end{figure}

\begin{table}[t]
\centering
\small
\caption{Autoregressive image generation acceleration on Infinity with synchronized timing. Time is reported in seconds per image.}
\label{tab:infinity-app}
\begin{tabular}{lcccc}
\toprule
Method & Time$\downarrow$ & Speedup$\uparrow$ & CLIPScore$\uparrow$ & ImageReward$\uparrow$ \\
\midrule
Original & 3.2511 & 1.00$\times$ & 0.3204 & 1.4110 \\
\method & 2.0791 & 1.56$\times$ & 0.3202 & 1.3133 \\
\bottomrule
\end{tabular}
\end{table}

\section{Additional Visual Comparisons}
\label{sec:more-visual}

\begin{figure*}[t]
  \centering
  \includegraphics[width=\textwidth,page=1]{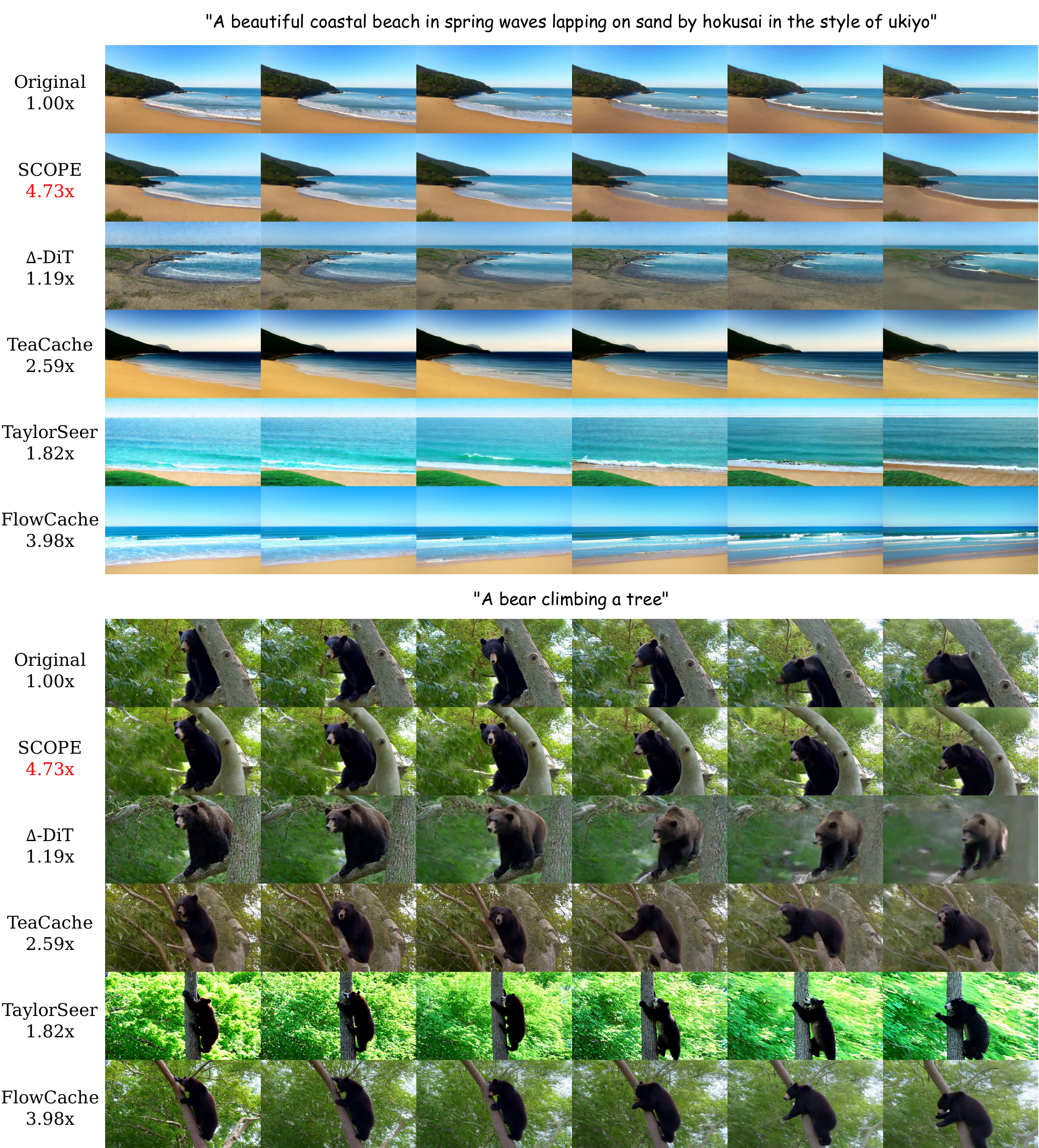}
  \caption{More Visual Comparisons on SkyReels-V2. Zoom in for best viewing.}
  \Description{Additional qualitative comparison on SkyReels-V2 across all methods.}
  \label{fig:sky_more}
\end{figure*}

\begin{figure*}[t]
  \centering
  \includegraphics[width=0.93\textwidth,page=1]{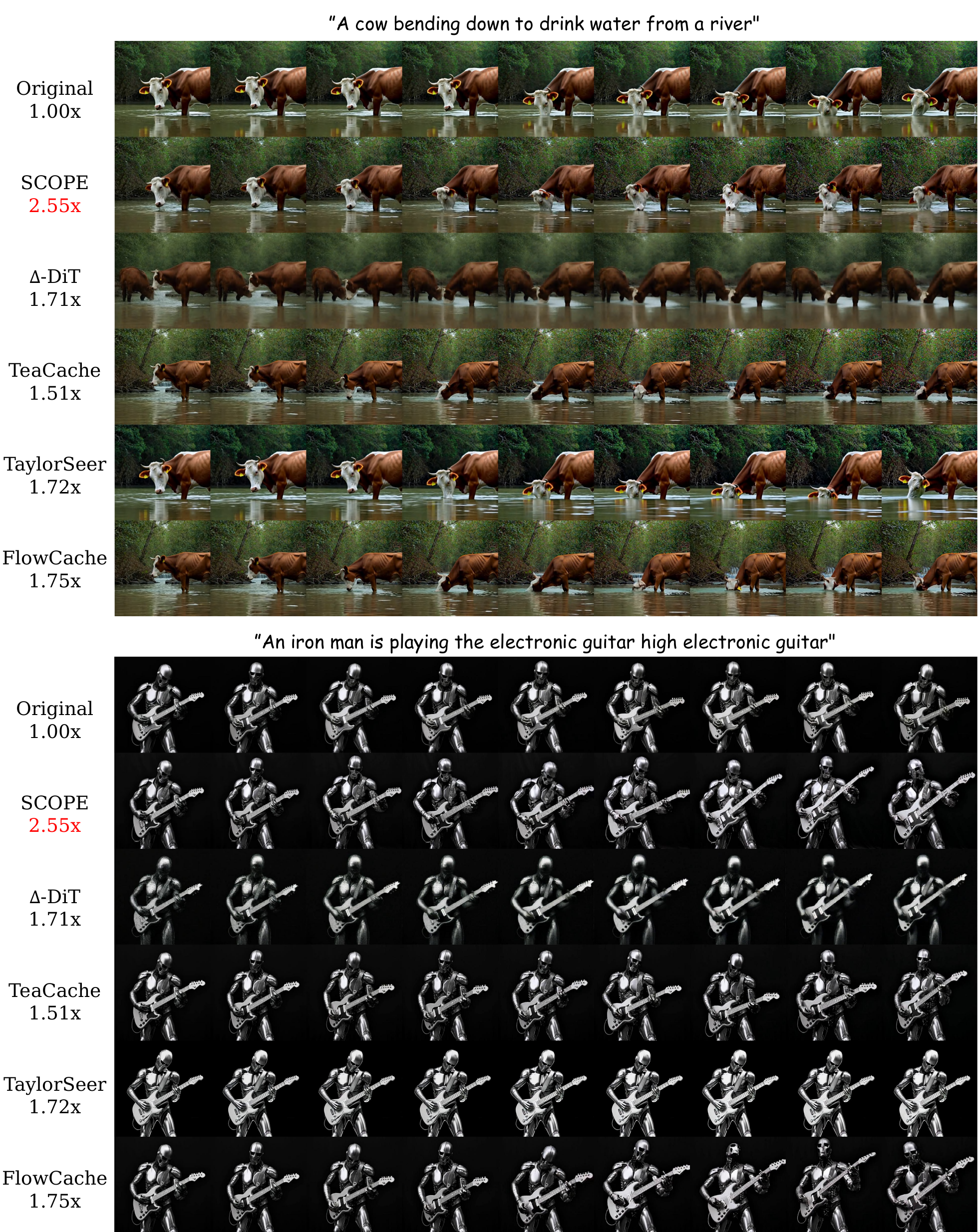}
  \caption{More Visual Comparisons on MAGI-1. Zoom in for best viewing.}
  \Description{Additional qualitative comparison on MAGI-1 across all methods.}
  \label{fig:magi_more}
\end{figure*}

We provide additional visual comparisons across SkyReels-V2 and MAGI-1 in figures~\ref{fig:sky_more} and~\ref{fig:magi_more}. The comparisons cover the same method set as the main table, including Original, \deltadit, TeaCache, TaylorSeer, FlowCache, and \method.
%  For each model, we pair runtime-oriented plots with representative generated frames or video strips so that the reader can directly inspect how acceleration differences relate to temporal stability, structural consistency, and detail preservation.

\end{document}